%% file: root.tex
\documentclass[letterpaper,10pt,journal,twoside]{IEEEtran}

\IEEEoverridecommandlockouts

\usepackage[caption=false, font=footnotesize]{subfig}

\usepackage[acronym]{glossaries}
\usepackage{booktabs}
\usepackage{textcomp}

\usepackage{graphicx}
\usepackage{svg}
\usepackage{subcaption}   
\usepackage{xcolor}
\usepackage{epsfig} 
\usepackage{graphics} 
\usepackage{times} 
\usepackage{amsmath,amsfonts} 
\usepackage{booktabs, multirow}
\usepackage{soul}
\usepackage{changepage,threeparttable} 
\usepackage{amssymb}  
\usepackage{graphicx}
\usepackage{lipsum}  
\usepackage{amssymb}
\usepackage{array}
\usepackage{flushend}
\usepackage[nosort,noadjust]{cite}
\usepackage{amsmath}

\usepackage{float}
\usepackage{microtype}
\usepackage{xargs}
\usepackage{booktabs}
\usepackage{bm}
\usepackage{dsfont}

\usepackage[hidelinks]{hyperref}
\usepackage[capitalise]{cleveref}
\usepackage{float}

\definecolor{myred}{HTML}{ff3030}

\definecolor{IEEEblue}{RGB}{0,102,161}

\title{
An Experimental Study of Downwash Effects on a Continuum Manipulator Integrated with a Multirotor UAV}
\author{Anuraj Uthayasooriyan,
Krishna Manaswi Digumarti,
Fernando Vanegas,
and Felipe Gonzalez %
\thanks{Manuscript received: January 14, 2026; Revised: April 10, 2026; Accepted: June 3, 2026. This paper was recommended for publication by Editor Yong-Lae Park upon evaluation of the Associate Editor and Reviewers' comments.}%
\thanks{The authors wish to acknowledge the continued support from the Queensland University of Technology (QUT) through the QUT Centre for Robotics, the support of the Research Engineering Facility (REF) team  and the senior technicians  Steven Bulmer, Benjamin Brownlee, and Amir Moghaddam at QUT for the provision of expertise and research infrastructure, and the help from Julian Galvez (REF), and PhD researcher Dipraj Debnath.}%
\thanks{The authors are with the QUT Centre for Robotics, School of Electrical Engineering and Robotics, QUT, Brisbane, QLD 4000, Australia.}%
\thanks{Corresponding author: Anuraj Uthayasooriyan, {\tt\footnotesize uanuraj@hotmail.com, uthayaso@qut.edu.au}.}%
\thanks{Digital Object Identifier (DOI): see top of this page.}%
\vspace{-0.5cm}
}

\input{notation}

\usepackage{eso-pic}
\usepackage{url}
\AddToShipoutPicture*{%
     \AtTextUpperLeft{%
         \put(-3.5,10){
           \begin{minipage}{\textwidth
           }
              \scriptsize
              \MakeUppercase{Preprint version of an IEEE Robotics and Automation Letters article; final version available at} \url{https://doi.org/10.1109/LRA.2026.3709580}
           \end{minipage}}%
     }%
}

\begin{document}

\maketitle

\input{0-abstract}
\input{1-introduction}

\input{2-DesignAndArchitecture}

\input{3-Bench_Experiment}

\input{4-Results_and_Flight}

\input{5-Discussion}
\input{6-Conclusions}
\bibliographystyle{IEEEtran} 
\bibliography{references}

\end{document}

%% file: notation.tex
\usepackage{amsmath}
\usepackage{amssymb}
\usepackage{accents}
\usepackage{bm}
\usepackage{xifthen}
\usepackage{color}
\usepackage{fancyvrb}
\usepackage{graphicx,scalerel}

\newcommand{\ba}{\begin{eqnarray}}
\newcommand{\ea}{\end{eqnarray}}

\newcommand{\presup}[1]{\,{}^{\scriptscriptstyle #1}\!}

\newcommand{\pose}[1][ZZZZ]{\ifthenelse{\equal{#1}{ZZZZ}}{}{\presup{#1}}{\mathbf{\xi}}}
\newcommand{\estpose}[1][ZZZZ]{\ifthenelse{\equal{#1}{ZZZZ}}{}{\presup{#1}}{\mathbf{\hat{\xi}}}}
\newcommand{\hpose}[1][ZZZZ]{\ifthenelse{\equal{#1}{ZZZZ}}{}{\presup{#1}}{\hat{\mathbf{\xi}}}}
\newcommand{\posedot}[1][ZZZZ]{\ifthenelse{\equal{#1}{ZZZZ}}{}{\presup{#1}}{\mathbf{\nu}}}

\newcommand{\q}[1][ZZZZ]{\ifthenelse{\equal{#1}{ZZZZ}}{}{\presup{#1}}{\mathring{q}}}

\DeclareMathAlphabet{\mathitbf}{OML}{cmm}{b}{it}
\newcommand{\twist}[2][ZZZZ]{\ifthenelse{\equal{#1}{ZZZZ}}{}{\presup{#1}}{\mathcal{S}}}
\renewcommand{\vec}[2][ZZZZ]{\ifthenelse{\equal{#1}{ZZZZ}}{}{\presup{#1}}{\mathitbf{#2}}}

\newcommand{\hvec}[2][ZZZZ]{\ifthenelse{\equal{#1}{ZZZZ}}{}{\presup{#1}}{\tilde{\vec{#2}}}}
\newcommand{\obvec}[2][ZZZZ]{\ifthenelse{\equal{#1}{ZZZZ}}{}{\presup{#1}}\rlap{${\overbridge{\phantom{$\vec{#2}$}}}$}\vec{#2}}
\newcommand{\evec}[2][ZZZZ]{\ifthenelse{\equal{#1}{ZZZZ}}{}{\presup{#1}}{\hat{\vec{#2}}}}
\newcommand{\bvec}[2][ZZZZ]{\ifthenelse{\equal{#1}{ZZZZ}}{}{\presup{#1}}{\bar{\vec{#2}}}}

\newcommand{\dvec}[2][ZZZZ]{\ifthenelse{\equal{#1}{ZZZZ}}{}{\presup{#1}}{\dot{\vec{#2}}}}
\newcommand{\ddvec}[2][ZZZZ]{\ifthenelse{\equal{#1}{ZZZZ}}{}{\presup{#1}}{\ddot{\vec{#2}}}}

\newcommand{\mat}[2][ZZZZ]{\ifthenelse{\equal{#1}{ZZZZ}}{}{\presup{#1}\,}{{\boldsymbol #2}}}
\newcommand{\dmat}[2][ZZZZ]{\ifthenelse{\equal{#1}{ZZZZ}}{}{\presup{#1}\,}{{\dot{\boldsymbol #2}}}}
\newcommand{\emat}[2][ZZZZ]{\ifthenelse{\equal{#1}{ZZZZ}}{}{\presup{#1}\,}{\hat{\boldsymbol#2}}}
\newcommand{\matfn}[3][ZZZZ]{\ifthenelse{\equal{#1}{ZZZZ}}{}{\presup{#1}}{{\mat{#2}}\left(#3\right)}}
\newcommand{\Rt}[2][ZZZZ]{\ifthenelse{\equal{#1}{ZZZZ}}{}{\presup{#1}}{{\bf R}\left(#2\right)}}

\newcommand{\point}[2][ZZZZ]{\ifthenelse{\equal{#1}{ZZZZ}}{}{\presup{#1}}{\mathbf{\mathrm{#2}}}}

\newfont{\School}{pncr}
\newfont{\eightTR}{pncr at 8pt}

\usepackage{color}
\usepackage{fancyvrb}
\fvset{formatcom=\color{blue},fontseries=c,fontfamily=courier,xleftmargin=4mm,commentchar=!}

\DefineVerbatimEnvironment{Code}{Verbatim}{formatcom=\color{blue},fontseries=c,fontfamily=courier,fontsize=\footnotesize,xleftmargin=4mm,commentchar=!}

\DefineVerbatimEnvironment{CodeSmall}{Verbatim}{formatcom=\color{blue},fontseries=c,fontfamily=courier,fontsize=\scriptsize,xleftmargin=1mm,commentchar=!}

\DefineVerbatimEnvironment{CodeNum}{Verbatim}{numbers=left,numbersep=4pt,formatcom=\color{blue},fontseries=c,fontfamily=courier,fontsize=\footnotesize,xleftmargin=4mm}

\newcommand{\model}[1]{\index{code}{#1@\textit{#1}}\ifthenelse{\boolean{draft}}{{\color{green}\Verb+#1+}}{\Verb+#1+}}
\newcommand{\block}[1]{\ifthenelse{\boolean{draft}}{{\color{green}\Verb+#1+}}{\textsf{#1}}}

\newcommand{\func}[2][ZZZZ]{\ifthenelse{\equal{#1}{ZZZZ}}{\index{code}{#2}}{\index{code}{#1}}\ifthenelse{\boolean{draft}}{{\color{green}\Verb+#2+}}{\Verb+#2+}}

\newcommand{\methodb}[2]{\index{code}{#1@\textbf{#1}!.#2}\ifthenelse{\boolean{draft}}{{\color{magenta}\Verb+#1.#2+}}{\Verb+#1.#2+}}
\newcommand{\method}[2]{\index{code}{#1@\textbf{#1}!.#2}\ifthenelse{\boolean{draft}}{{\color{magenta}\Verb+#2+}}{\Verb+#2+}}
\newcommand{\class}[1]{\index{code}{#1@\textbf{#1}}\ifthenelse{\boolean{draft}}{{\color{cyan}\Verb+#1+}}{\Verb+#1+}}
\newcommand{\property}[1]{\index{property}{#1}\ifthenelse{\boolean{draft}}{{\color{cyan}\Verb+#1+}}{\Verb+#1+}}



%% file: 0-abstract.tex
\begin{abstract}

Continuum arm aerial manipulation systems leverage soft-manipulator compliance and dexterity for tasks in confined or hazardous environments, but propeller downwash can degrade performance, particularly near walls and the ground. This effect remains uncharacterized for continuum manipulators. This letter experimentally studies downwash-induced kinematic deviations of a tendon-driven continuum manipulator integrated with a multirotor platform. Under still-air conditions, the CM is compared with a constant-curvature (CC) model. Downwash-induced end-effector pose deviations are then quantified relative to the mean still-air experimental baseline at four propeller throttle levels in free space, and at maximum throttle near a wall, and near the ground. Vertical position and yaw show the largest deviations and are amplified by ground effect. A CC-guided Gaussian process regression (GPR) residual model is learned from experimental data that improves forward pose prediction RMSE (position by 89-95$\%$, orientation by 47-79$\%$), and support compensation-oriented, downwash-aware modeling of continuum arm aerial manipulation systems.
\end{abstract}


%% file: 1-introduction.tex
\vspace{-0.4cm}
\section{Introduction}
 \IEEEPARstart{A}{erial} manipulators are multi-rotor platforms designed to perform tasks, using devices attached to their bodies.They are especially useful in hard-to-reach, risky, or low-visibility environments \cite{ollero2021past}. Their use spans industrial inspections\cite{Hamaza2020}, carrying payloads, agricultural, and environmental sample collection \cite{ollero2021past}. Current research is focused on two main types of manipulators. Conventional designs use articulated manipulators or grippers \cite{ollero2021past}, whereas recent designs use continuum manipulators (CM) or soft robotic grippers \cite{uthayasooriyan2024tendon,peng2025dexterous,Ubellacker2024}. 
 Despite their high load capacity, rigid-link arms struggle with dexterous, compliant motion along tortuous paths, which restricts their use in industrial inspection or environmental sampling. Continuum arm aerial manipulation systems (CAAMS) are comparatively less heavy and demonstrate high dexterity, compliance and maneuverability \cite{peng2025dexterous,jalali2022aerial}.

Because of its soft structure, the CM is highly susceptible to downwash (DW) (downward movement of air) from propellers. This is particularly relevant when operating near walls, frames, or the ground during inspection, maintenance, or sampling tasks. In these close-proximity regimes, downwash–surface interaction  can intensify turbulence\cite{david2024ground,WU2024204, samadikhoshkho2020modeling}, degrading manipulator stability and end-effector (EE) accuracy. The study of CAAMS is relatively new and largely split between mechanical design and, dynamics and control. Control studies, model DW as a sinusoidal disturbance. There is thus a lack of experimental study and the incorporation of real-world data into mathematical models. On the design side, researchers have proposed tendon-driven continuum arms \cite{zhao2022modular}, ground-based tendon actuation for tethered CAAMS \cite{chien2023design,chien2021kinematic}, design and control of three-section continuum manipulators \cite{peng2023aecom,peng2025dexterous}, and eye--in--hand visual servoing for EE control \cite{amiri2025high}. The importance of disturbance estimation, compensation, and mitigation in aerial robots is evident in  \cite{khalifa2024sensorless,lee2022towards,wei2025passive}. However, DW effects remain largely unexamined in the current CAAMS literature \cite{samadikhoshkho2020modeling,hashemi2023robust,ghorbani2023dual}.

{%
\begin{figure}[t]
    \setlength{\abovecaptionskip}{-1pt}
    
    \centering
    \includegraphics[width=\linewidth]{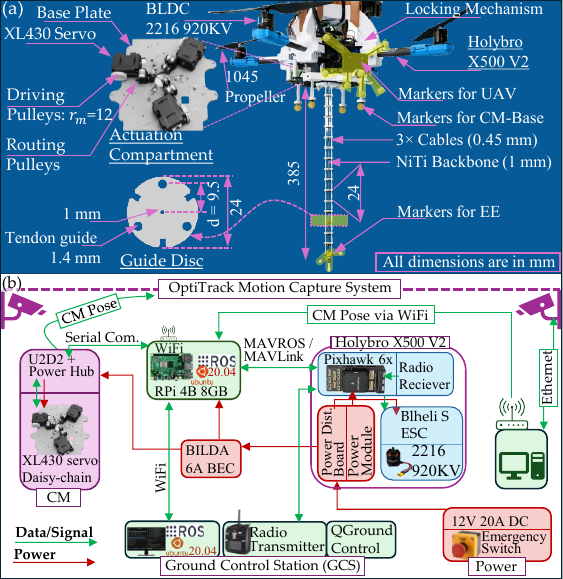}
    \caption{(a) The mechanical  and (b) electrical system architectures of the CAAMS used for the experimental study.  }
    \label{fig:System_Architecture}
    \vspace*{-0.6cm}
\end{figure}
}%

\begin{figure*}[!t]
    \vspace{4pt}
    \centering
    \includegraphics[width=\linewidth]{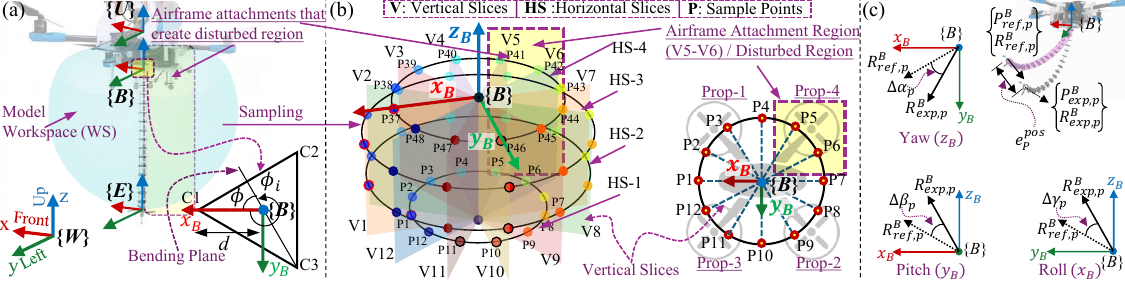}
    \caption{(a) CAAMS with the CC workspace and frames: world \(\{W\}\), UAV \(\{U\}\), CM base \(\{B\}\), and CM tip \(\{E\}\). The inset triangle marks cables C1--C3. (b) Workspace sampling: the 3D workspace is divided into four horizontal slices (HS1--HS4) and 12 vertical slices (V1--V12) about \(Z_B\), whose intersections define points \(P1\)–\(P48\). The circular panel shows HS1 in top view with dashed vertical slices, propellers and Dashed-yellow shaded boxes: Airframe attachments causing flow disturbance. (c) Definition of position and orientation error between the expected \(\{ref\}\) and measured \(\{exp\}\) pose in \(\{B\}\).}
    \label{fig:workspace}
    \vspace{-10pt}
\end{figure*}
In this work, we experimentally characterize DW-induced pose deviations of a tendon-driven continuum manipulator on a multirotor UAV, together with a data-driven correction model. Under still-air conditions, the manipulator is benchmarked against a constant-curvature (CC) model. Keeping the  mean still-air experimental results as baseline, the DW-induced deviations are relatively evaluated. In parallel, a CC-guided Gaussian process regression (GPR) residual model is learned to improve forward pose prediction under still air and the DW. The contributions of this work are an experimental evaluation of DW effect from a multi-rotor UAV on:

\begin{enumerate}
    \item  An airborne continuum manipulator's end effector pose accuracy when the system is in free-space and the additive torque variation of driving motors.
    \item An airborne continuum manipulator's end effector pose accuracy in the proximity of a wall and  ground.
    \item A CC-guided GPR residual model that compensates CC-to-prototype mismatch and improves forward pose prediction in still-air and downwash conditions. 
\end{enumerate}

\vspace{-0.4cm}

%% file: 2-DesignAndArchitecture.tex
\section{Design and Architecture}
\label{sec:System_Architecture}

\subsection{System architecture} 

\cref{fig:System_Architecture} depicts the evaluated CAAMS: \ref{fig:System_Architecture}(a) shows the mechanical architecture, combining a Holybro X500 V2 quadrotor with a tendon-driven CM, and (b) the electrical architecture. A ground control station (GCS) communicates with an onboard Raspberry Pi 4B to send tendon-length commands to the CM's EE. Both run ROS Noetic on Ubuntu 20.04, and QGroundControl provides flight-controller access.
\vspace{-0.4cm}

\subsection{Continuum manipulator} 
We adopt a single-section CM design inspired by \cite{GrassmannBurgner-Kahrs_et_al_Frontiers_2023,5957337}. Its dimensions are given in \cref{fig:System_Architecture}(a), with a segment-length-to-tendon-guide-spacing ratio = 0.4, to improve CC approximation accuracy \cite{uthayasooriyan2024tendon, li2002design}. The CM uses three braided microfilament thread tendons \cite{amanov2021tendon}, spaced at 120$^\circ$ around a central backbone (nitinol (NEXMETAL)) and driven by three Dynamixel XL430-W250-T servos with pulleys of radius \(r_m=12\) mm, reducing actuator count and mass compared with a four-tendon layout \cite{uthayasooriyan2024tendon,webster2010design}. The guide disks are 3D-printed PLA bonded with epoxy adhesive.


\subsection{Kinematic model} 
\label{subsec:Kinematic model}

The CM's CC model follows \cite{webster2010design}, with axis orientations and cable alignment as in \cref{fig:workspace}(a). The frames are \(\{W\}\) (world), \(\{U\}\) (UAV), \(\{B\}\) (CM base), and \(\{E\}\) (CM tip). The \(x\)-axis of \(\{B\}\) aligns with cable 1 at the base, the CM extends along \(-z_B\), and \(\{E\}\) is aligned with \(\{B\}\) in the rest configuration. The EE pose in \(\{W\}\) is given by \(\mathbf{T}_{WE} = \mathbf{T}_{WU}\mathbf{T}_{UB}\mathbf{T}_{BE}\), where \(\mathbf{T}_{WU}\) is estimated using the motion-capture system in \cref{fig:System_Architecture}(b), and \(\mathbf{T}_{UB}\) is fixed by design. The constant curvature model of the CM with an inextensible backbone is defined by curvature $\kappa$, bending plane angle $\phi$ measured from positive x--axis and length of the backbone $\ell$. Defining $c_\phi = \cos\phi$, $s_\phi = \sin\phi$, $c_{\kappa\ell} = \cos({\kappa\ell})$, and $s_{\kappa\ell} = \sin({\kappa\ell})$, the pose of the EE in frame $\{B\}$ is

{%
\setlength{\abovedisplayskip}{-6pt}%
\begin{equation}
\label{eq:CM_homogeneous}
T_{BE} =
\begin{bmatrix}
c_{\kappa\ell} c_\phi & s_\phi & s_{\kappa\ell} c_\phi & \frac{1}{\kappa}(1-c_{\kappa\ell})c_\phi \\
c_{\kappa\ell} s_\phi & -c_\phi & s_{\kappa\ell} s_\phi & \frac{1}{\kappa}(1-c_{\kappa\ell})s_\phi \\
s_{\kappa\ell} & 0 & -c_{\kappa\ell} & -\frac{1}{\kappa}s_{\kappa\ell} \\
0 & 0 & 0 & 1
\end{bmatrix}.
\end{equation}
}%

For a symmetric layout with cable offset \(d\) and backbone length \(L\), the CM kinematic parameters are functions of the cable-length actuation vector \(q=[l_1,l_2,l_3]^T\):
{\setlength{\abovedisplayskip}{4pt}
\setlength{\belowdisplayskip}{4pt}
\setlength{\jot}{2pt}
\begin{align}
\ell(q) &= L \label{l_q}\\
\phi(q) &= \operatorname{atan2}\!\left(\sqrt{3}(l_2-l_3),\, l_2+l_3-2l_1\right) \label{BendingPlaneAngle}\\
\kappa(q) &= \frac{2}{d(l_1+l_2+l_3)}
\sqrt{l_1^2+l_2^2+l_3^2-l_1l_2-l_1l_3-l_2l_3}
\label{Kappa}
\end{align}}
Here, \(l_i\) is the length of cable \(i\), \(i=1,2,3\). The tendon length change is \(\Delta l_i=-L\kappa d\cos(\phi_i)\), where \(\phi_i\) is the angular position of cable \(i\) at the base relative to the bending plane. For a pulley of radius \(r_{m_i}\), a motor rotation \(\psi_i\) gives \(\Delta l_i=r_{m_i}\psi_i\).

%% file: 3-Bench_Experiment.tex
\section{Experiments} 
\label{sec:Bench_Experiments}

\subsection{Experiment Design}


{%
\begin{figure}[htb]
    \vspace{4pt}
    \setlength{\abovecaptionskip}{-1pt}
    \setlength{\belowcaptionskip}{-4pt}
    \centering
    \includegraphics[width = \linewidth]{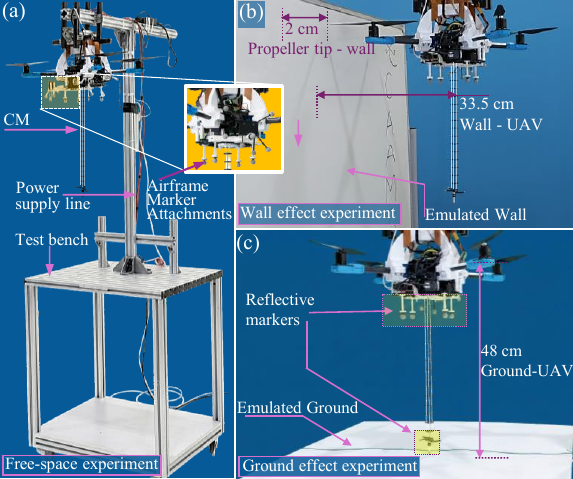}
    \caption{ Bench-top experiment setups for (a)free-space (b) wall effect where a board is placed next to the CAAMS, and (c) ground effect where the UAV is close to a horizontal board.}
    \label{fig:ExperimentSetup}
    \vspace{-0.8\baselineskip}
\end{figure}
}%

We designed a bench--top study to benchmark the CM EE pose against its CC model without DW in Case (i), then evaluate three DW cases: free-space (ii), near-wall (iii), and near-ground (iv), as given in \cref{tab:experimentDesign}. Case (ii) used four throttle levels as repeatable operating settings monitored in QGroundControl rather than absolute RPM control. Cases (iii) and (iv) were tested only at \(T100\), assuming the worst-case condition for wall/ground proximity effects.

%

\subsection{Experiment Setup}

We mounted the CAAMS onto the test bench shown in \cref{fig:ExperimentSetup} while completely lock all 6DOF. A 12 V, 20 A DC supply powers the CAAMS. OptiTrack\text{\textregistered} motion-capture system tracks frames \(\{U\}\), \(\{B\}\), and \(\{E\}\) using the reflective markers. We used the configurations in \cref{fig:ExperimentSetup}(a) for Cases~(i)–(ii), \cref{fig:ExperimentSetup}(b) for Case~(iii), and \cref{fig:ExperimentSetup}(c) for Case~(iv). For wall-effect tests, a flat board was placed parallel to the front rotors (Prop~1 and Prop~2 in \cref{fig:workspace}(b)) at 2 cm beyond the rotor tip radius for a closer but safe operating condition. For ground-effect, a parallel horizontal board was placed beneath the UAV, with the CM tip just touching the surface and the board 48 cm below the propeller-arm plane (sampling/picking case). The manipulator is centrally mounted beneath the UAV hub, with arm-length-to-wheelbase ratio \(385/500=0.77\) and propeller-plane-to-EE distance \(\approx 490~\mathrm{mm}\). Using 1045 propellers \((R\approx127~\mathrm{mm})\), the 525 mm board-to-propeller-plane distance gives \(h/R\approx4.13\), indicating not an extreme classical ground-effect condition \cite{WU2024204}, but a near-ground DW--surface interaction.
\vspace{-0.2cm}


\subsection{Data Collection}
\vspace{-0.7\baselineskip}
{%
\begin{table}[h]
    \centering
    \caption{Bench-top experiments under different cases}
    \label{tab:experimentDesign}
    \vspace{-0.6\baselineskip}
    \setlength{\tabcolsep}{2pt}  
    \resizebox{\columnwidth}{!}{%
    \begin{tabular}{>{\raggedright\arraybackslash}p{0.15\linewidth}>{\raggedright\arraybackslash}p{0.6\linewidth}>{\raggedright\arraybackslash}p{0.25\linewidth}}
        \toprule
        Experiment & Physical interpretation& Throttle levels T\%\\
        \midrule
        Case(i)& No DW& 0 \\
        Case(ii)& DW;  No influence of wall/ground& 25, 50, 75, 100\\
        Case(iii)& DW under the influence of wall& 100 \\
        Case(iv)& DW under the influence of ground& 100 \\
        \bottomrule
    \end{tabular}}
    
\end{table}

\vspace{-1.0\baselineskip}
}%

We first sampled the theoretical workspace (WS) in \cref{fig:workspace}(a), generated by \cref{eq:CM_homogeneous}, restricting the bending angle to \(\leq 180^\circ\) to avoid propeller interference. As shown in \cref{fig:workspace}(b), the WS was divided into 4--horizontal slices (HS1–HS4) and 12 vertical slices (V1–V12) with \(\phi=30^\circ\), giving 48 sample points. Using the CC model (\cref{subsec:Kinematic model}), we precomputed the tendon lengths for these poses. The CAAMS was then remotely actuated to each pose, and for each case in \cref{tab:experimentDesign}, all CAAMS frame poses and actuation variables were recorded in ROS \texttt{.bag} files. At each sample pose, based on observations, the CM was held for 10 seconds to allow the CM to settle and recorded the steady-state pose for next 10 seconds. Each experiment was repeated three times.
\vspace{-0.3cm}


\subsection{Evaluation of data}
\vspace{-0.2cm} 

In \cref{sec:Results}, Case~(i) assesses the prototype against the CC model, while Cases~(ii)–(iv) use the still-air baseline to quantify DW-induced EE pose errors. For each sample at point \(P\), pose errors are expressed in \(\{B\}\) using the Euclidean position error \(e_{p}^{\mathrm{pos}}\) in \eqref{eq:pos_error} and scalar orientation error \(e_{R,p}^{\circ}\) in \eqref{eq:ori_error}. Here, \(e_{p}^{\mathrm{pos}}\) is the measured–position's Euclidean distance in mm, and \(e_{R,p}^{\circ}\) is the principal angle of the relative rotation \(\mathbf{R}_{\mathrm{rel},p}\). For Case~(i), the reference pose is from the CC model at the corresponding WS point. For Cases~(ii)–(iv), the reference is the mean repeated Case~(i) measurements at each WS point: positions are averaged linearly, orientations as unit quaternions using Markley’s method~\cite{markley2007quaternion}, and Euler components are obtained from \eqref{eq:euler_rel_zyx}.

\subsubsection{Position error}

As shown in \cref{fig:workspace}(b),(c), at every workspace point $P$,  let $\mathbf{P}^{B}_{\mathrm{exp}}=[x^{B}_{\mathrm{exp},p},\,y^{B}_{\mathrm{exp},p},\,z^{B}_{\mathrm{exp},p}]^{\top}$ denote the measured EE position for the sample, and $\mathbf{p}^{B}_{\mathrm{ref}}=[x^{B}_{\mathrm{ref},p},\,y^{B}_{\mathrm{ref},p},\,z^{B}_{\mathrm{ref},p}]^{\top}$ denote the corresponding reference EE position. With $\Delta x_{p}=x^{B}_{\mathrm{exp},p}-x^{B}_{\mathrm{ref},p}$, $\Delta y_{p}=y^{B}_{\mathrm{exp},p}-y^{B}_{\mathrm{ref},p}$, and $\Delta z_{p}=z^{B}_{\mathrm{exp},p}-z^{B}_{\mathrm{ref},p}$, the Euclidean position error is
{\setlength{\abovedisplayskip}{4pt}
\setlength{\belowdisplayskip}{4pt}
\begin{equation}
e_{p}^{\mathrm{pos}}
=
\sqrt{
\Delta x_{p}^{2}+\Delta y_{p}^{2}+\Delta z_{p}^{2}
},
\label{eq:pos_error}
\end{equation}
}

\subsubsection{Orientation error}

Let $\mathbf{R}^{B}_{\mathrm{exp},p}\in SO(3)$ and $\mathbf{R}^{B}_{\mathrm{ref},p}\in SO(3)$ be the measured and reference EE orientations at the same WS point. Following \cite{lynch2017modern}, the relative rotation is
{\setlength{\abovedisplayskip}{4pt}
\setlength{\belowdisplayskip}{4pt}
\begin{equation}
\mathbf{R}_{\mathrm{rel},p}
=
\left(\mathbf{R}^{B}_{\mathrm{exp},p}\right)^{-1}\mathbf{R}^{B}_{\mathrm{ref},p}.
\label{eq:rel_error}
\end{equation}
}

The scalar orientation error is defined as the principal rotation angle of $\mathbf{R}_{\mathrm{rel},p}$:
{\setlength{\abovedisplayskip}{4pt}
\setlength{\belowdisplayskip}{4pt}
\begin{equation}
e_{R,p}^{ori} = \angle\!\left(\mathbf{R}_{\mathrm{rel},p}\right)\in[0,\pi],
\qquad
e^{\circ}_{R,p}=\frac{180}{\pi}\,e_{R,p}^{ori},
\label{eq:ori_error}
\end{equation}
}

For qualitative interpretation, $\mathbf{R}_{\mathrm{rel},p}$ is decomposed into extrinsic (fixed-axis) ZYX Euler angles \cite{scipy_from_euler,scipy_as_euler}:
{\setlength{\abovedisplayskip}{4pt}
\setlength{\belowdisplayskip}{4pt}
\begin{equation}
\begin{bmatrix}
\Delta\alpha_{p} & \Delta\beta_{p} & \Delta\gamma_{p}
\end{bmatrix}^{\!\top}
=
\mathrm{Euler}_{\text{extrinsic }zyx}\!\left(\mathbf{R}_{\mathrm{rel},p}\right),
\label{eq:euler_rel_zyx}
\end{equation}
}

where $\Delta\alpha$, $\Delta\beta$, and $\Delta\gamma$ are the yaw--pitch--roll corrections (refer \cref{fig:workspace}(c)) about the fixed base-frame axes $z_B,y_B,x_B$, respectively. Each component is wrapped to $[-180^{\circ},180^{\circ}]$ to avoid angle discontinuities. Here, the yaw: azimuthal rotation about $z_B$, pitch: tilt about $y_B$, and roll: tilt about about $x_B$. 

\subsubsection{Mean and standard deviation (SD) used in plots}

For each workspace point $P$ with $N_{p}$ and the samples $k$, the plotted mean and SD are computed from the set of available scalar errors $\{e_{p,k}\}_{k=1}^{N_{p}}$ (either $e_{p}^{\mathrm{pos}}$ in \eqref{eq:pos_error} or $e^{\circ}_{R,p}$ in \eqref{eq:ori_error}), with number of samples pooled for point $p$ as
{\setlength{\abovedisplayskip}{4pt}
\setlength{\belowdisplayskip}{-4pt}
\setlength{\jot}{2pt}
\begin{align}
\mu_{p} &= \frac{1}{N_{p}}\sum_{k=1}^{N_{p}} e_{p,k}, \label{eq:mean}
\end{align}

\begin{align}
\sigma_{p} &= \sqrt{\frac{1}{N_{p}}\sum_{k=1}^{N_{p}}\left(e_{p,k}-\mu_{p}\right)^{2}}
\label{eq:std}
\end{align}

}


\subsection{CC-guided residual prediction from experimental data}

\label{sec:gpr_residual_model}

The large mismatch between the as-built manipulator and the ideal CC model motivates an empirical correction model. GPR has been used for tendon-actuated soft-arm kinematics \cite{RELANO2023107174}, residual robot calibration \cite{dacs2023active}, and uncertainty-aware continuum-robot state estimation \cite{lilge2022continuum}, supporting its use here as a physics-guided residual model. Our approach learns a correction to a CC prior over a reduced actuation space and evaluates end-effector pose prediction under still-air and DW. Here, rather than replacing the CC kinematics, GPR learns the position and orientation residuals from experimental data, providing a predictive, compensation-oriented model while retaining the nominal CC prior.
The DW-induced errors reported in \Cref{subsec: T0 vs T_ON} to IV-D remain defined relative to the mean still-air experimental baseline; the GPR model is used separately for forward prediction based on the CC prior.

For each sample, one representative pose is computed from the measured data. Using the reduced actuation coordinates as \(a=l_2+l_3-2l_1\) and \(b=\sqrt{3}(l_2-l_3)\), and input \(\mathbf{x}=[a,b]^\top\), the CC workspace points are converted into continuous interpolants returning the nominal pose \((\hat{\mathbf{p}}_{CC}(\mathbf{x}),\hat{\mathbf{R}}_{CC}(\mathbf{x}))\). The residual targets are \(\mathbf{r}_p=\mathbf{p}_{exp}-\hat{\mathbf{p}}_{CC}\) and \(\mathbf{r}_r=\log_{SO(3)}(\hat{\mathbf{R}}_{CC}^{\top}\mathbf{R}_{exp})\), with \(\mathbf{r}_p=[r_{px},r_{py},r_{pz}]^\top\) and \(\mathbf{r}_r=[r_{rx},r_{ry},r_{rz}]^\top\). Six scalar GPR regressors are fitted to \([r_{px},r_{py},r_{pz},r_{rx},r_{ry},r_{rz}]\) using MATLAB \texttt{fitrgp}. Upon initial trials, hyperparameters are selected by grouped four-fold cross-validation over unique training points using
{\setlength{\abovedisplayskip}{3pt}
\setlength{\belowdisplayskip}{3pt}
\begin{equation}
J=
\frac{\mathrm{RMSE}^{GPR}_{pos}}{\mathrm{RMSE}^{CC}_{pos}}
+
\frac{\mathrm{RMSE}^{GPR}_{ori}}{\mathrm{RMSE}^{CC}_{ori}},
\end{equation}} 
Smaller \(J\) indicates larger simultaneous improvement in position and orientation.  \(\mathrm{RMSE}^{GPR}\) and \(\mathrm{RMSE}^{CC}\) are the RMSEs of the GPR and CC model relative to experiment. With \(\hat{\boldsymbol\mu}_p\) and \(\hat{\boldsymbol\mu}_R\) the GP posterior mean residuals, the corrected predictor is
{\setlength{\abovedisplayskip}{3pt}
\setlength{\belowdisplayskip}{3pt}
\begin{equation}
\hat{\mathbf p}=\hat{\mathbf p}_{CC}+\hat{\boldsymbol\mu}_p,
\qquad
\hat{\mathbf R}=\hat{\mathbf R}_{CC}\exp\!\left(\hat{\boldsymbol\mu}_R\right).
\end{equation}
}
We avoided the unstable region: V5--V6 is excluded due to attachment-disturbed fluctuations and HS4 due to large deviations, leaving HS1--HS3 for fitting. Thus, the usable workspace is limited to HS3, which is acceptable here as the aim is to assess GPR's feasibility. Points \(\{P2,P9,P16,P22,P31,P35\}\), covering different locations of the WS are held out for testing and the remaining are used for training. Initial trials restrict the kernel pool to \texttt{ardrationalquadratic}, \texttt{ardsquaredexponential}, and \texttt{ardmatern52} with constant basis; \texttt{ardrationalquadratic} is selected for all five cases. One model is trained per case \((T0,T25,T50,T75,T100)\), yielding a still-air calibration model at \(T0\) and DW-aware predictors for \(T25\)–\(T100\).

%% file: 4-Results_and_Flight.tex
\begin{figure}[!t]
    \vspace*{0.2cm}
    \centering
    \setlength{\abovecaptionskip}{-2pt}
    \setlength{\belowcaptionskip}{-8pt}
    \includegraphics[width=0.97\linewidth]{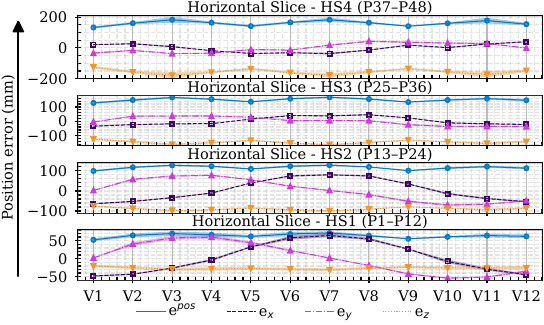}
    \caption{Case~(i) -- Euclidean (\(e_{pos}\)) and Cartesian EE position error components across slices with propellers off (\(T=0\%\))}
    \label{fig:IndivPosErrorT0}
    \vspace{-4pt}
\end{figure}

\begin{figure}[h]
    \centering
    \setlength{\abovecaptionskip}{-2pt}
    \setlength{\belowcaptionskip}{-8pt}
    \includegraphics[width=0.97\linewidth]{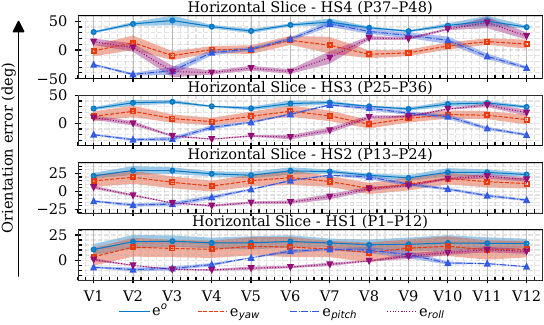}
    \caption{Case~(i) -- scalar quaternion errors (${e_{ori}}$) and the extrinsic Euler components of EE orientation with ($T=0\%$) }
    \label{fig:IndivOriErrorT0}
    \vspace{-6pt}
\end{figure}
\section{Results}
\label{sec:Results}

\subsection{Evaluating the CM prototype Vs CC model at ($T=0\%$)}
\label{sec:proto vs CC}


We first assess the CM prototype against the CC model at \(T0\). This baseline captures EE deviations tied to fabrication and modelling simplifications, and serves as the reference for later DW experiments. Since the goal is to study DW effects rather than optimize CM accuracy, the prototype is used as built. The GPR model is introduced separately to strengthen modelling baseline for forward prediction, while the DW-induced errors in Cases~(ii)--(iv) remain referenced to the mean still-air experimental baseline so that calibration residuals do not obscure the aerodynamic effect.

\subsubsection{Position Error of the Continuum Manipulator}

\begin{figure*}[!ht]
    \vspace*{0.1cm}
    \setlength{\abovecaptionskip}{-1pt}
    \centering
    \includegraphics[width=0.32\textwidth]{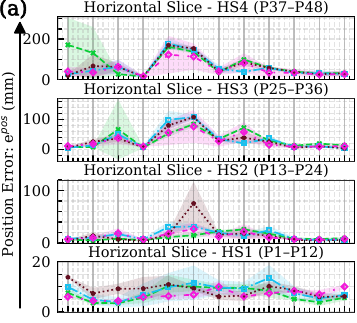}\hfill
    \includegraphics[width=0.32\textwidth]{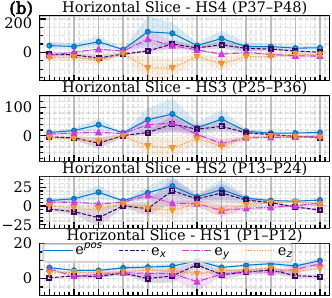}\hfill
    \includegraphics[width=0.32\textwidth]{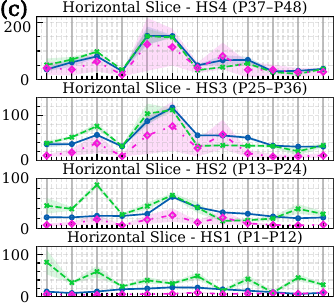}

    \vspace{2mm}

    \includegraphics[width=0.32\textwidth]{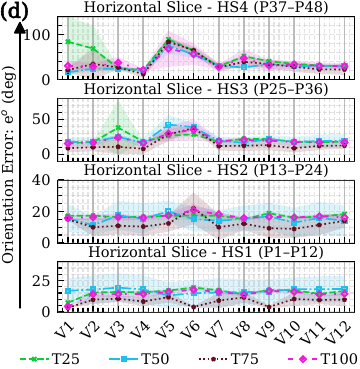}\hfill
    \includegraphics[width=0.32\textwidth]{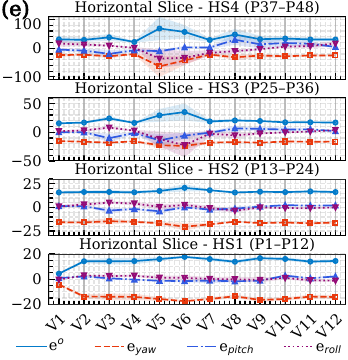}\hfill
    \includegraphics[width=0.32\textwidth]{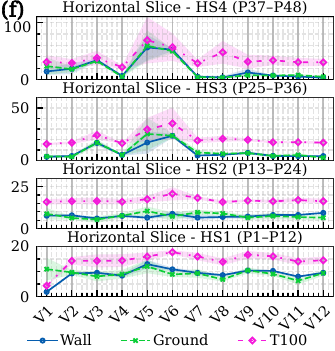}

    \caption[ ]%
    {EE errors of Euclidean position  \(e^{\mathrm{pos}}\) and scalar quaternion  orientation $e^{\circ}$ relative to mean at $T=0\%$ in Cases~(ii)--(iv).}
    \label{fig:combined_results}
    
    \vspace{1mm}
    \noindent
    \begin{tabular*}{\textwidth}{@{\extracolsep{\fill}}p{0.32\textwidth}p{0.32\textwidth}p{0.32\textwidth}@{}}
    \textbf{(a)} Case~(ii):  \(e^{\mathrm{pos}}\) across HS with DW at four throttle levels. &
    \textbf{(b)} Case~(ii): \(e^{\mathrm{pos}}\) and Cartesian  components across HS at $T=100\%$. &
    \textbf{(c)} Cases~(iii)$\&$(iv): \(e^{\mathrm{pos}}\) across HS \newline with wall and ground  at $T=100\%$. \\[1ex]

    \textbf{(d)} Case~(ii): $e^{\circ}$ across HS with DW at four throttle levels. &
    \textbf{(e)} Case~(ii):$e^{\circ}$ and extrinsic Euler components across HS at $T=100\%$. &
    \textbf{(f)} Cases~(iii)$\&$(iv):$e^{\circ}$ across HS\newline with wall and ground  at $T=100\%$. 
    \end{tabular*}
     \label{fig:combined_results}
    \vspace{-1.0\baselineskip}
\end{figure*}

\Cref{fig:IndivPosErrorT0} reports the Euclidean position error (\(e^{\mathrm{pos}}\)) between the prototype and the CC model over all sampled points in each horizontal slice (HS). The error remains bounded, with SD typically \(10\)–\(25~\mathrm{mm}\) and occasional peaks up to \(\approx35~\mathrm{mm}\) in HS4. The mean \(e^{\mathrm{pos}}\) increases with bending, from \(\approx50\)–\(75~\mathrm{mm}\) in HS1 to \(\approx150\)–\(200~\mathrm{mm}\) in HS4, with minima at V1, V5, and V9. Relative to the CM length (\(395~\mathrm{mm}\)), the maximum \(e^{\mathrm{pos}}\) increases from \(12.65\%\) in HS1 to \(50.6\%\) in HS4. In Cartesian components, \(e_z\) dominates and is consistently negative, shifting from \(\approx-30~\mathrm{mm}\) in HS1 to \(\approx-175~\mathrm{mm}\) in HS4 while remaining nearly constant within each slice. In contrast, \(e_x\) and \(e_y\) vary sinusoidally with V-index within roughly \(\pm50~\mathrm{mm}\). For HS1–HS3, \(e_x\) is positive over V5–V9 and negative over V10–V12 and V1–V4, while \(e_y\) is positive over V1–V6 and negative over V7–V12; this pattern reverses in HS4.

\subsubsection{Orientation Error of the Continuum Manipulator}

A similar trend is observed in the orientation error ($e^{\circ}$) (\cref{fig:IndivOriErrorT0}), with aligned minima and maxima. Its mean increases from \(\approx10\)–\(25^\circ\) in HS1 to \(\approx25\)–\(50^\circ\) in HS4, while SD remains \(\approx15^\circ\). Peaks coincide with indices where \(|e_{\mathrm{pitch}}|\) and/or \(|e_{\mathrm{roll}}|\) are largest. Both \(e_{\mathrm{pitch}}\) and \(e_{\mathrm{roll}}\) vary sinusoidally: \(e_{\mathrm{pitch}}\) is mostly negative for V1–V4 and V10–V12 and positive for V5–V10, whereas \(e_{\mathrm{roll}}\) is positive for V8–V12 and V1 and negative for V2–V7. By contrast, \(e_{\mathrm{yaw}}\) is smoother and largely positive, staying within \(\approx0\)–\(20^\circ\), while \(e_{\mathrm{pitch}}\) and \(e_{\mathrm{roll}}\) show larger excursions in HS3–HS4 (\(\approx-40^\circ\) to \(+40^\circ\))  with SD mostly $<10^{\circ}$.



\subsection{Evaluating the DW effect on the CM  in free-space}
\label{subsec: T0 vs T_ON}
Using the experiment mean values from \cref{sec:proto vs CC} as reference, we evaluate the DW effect on the CM by calculating the error in pose  at different throttle levels as in \cref{tab:experimentDesign}.

\subsubsection{Position error of the prototype under DW}
\label{subsubsec:pos_error_DW}


\cref{fig:combined_results}(a) shows that the DW-induced \(e^{\mathrm{pos}}\) is smallest in Horizontal Slice~1 (P13--P24), where the EE remains near its rest position, with mean of 5--15~mm and narrow SD. As it moves up through HS~2--4, both the mean and SD increase. In these top slices, peaks occur near V5--V6, where mean reach 25--180~mm, most prominently at T50 and T75, with large SD. Outside these regions, the \(e^{\mathrm{pos}}\) generally increases with throttle, while producing larger peaks and SD at T100.

\subsubsection{Orientation error of the prototype under DW}
\label{subsubsec:ori_error_DW}


The DW-induced orientation error ($e^{\circ}$) in \cref{fig:combined_results}(d) follows the \(e^{\mathrm{pos}}\) trends in \cref{fig:combined_results}(a). Relative to T0, HS~1 remains low (\(10^\circ\)–\(20^\circ\)) and varies smoothly with V-index. In HS~2–3, the mean increases and peaks at the same V-indices as the position error, with larger SD; T50 and T75 fluctuate most, while T100 is more stable despite comparable or slightly higher mean error. Slice~4 shows the strongest DW effect, with mean of \(60^\circ\)–\(100^\circ\) at V5 and V6, especially at low/intermediate throttles. In HS~1 and HS~4, T25 exceeds T75 or T100, indicating non-monotonic throttle dependence.
\vspace{-0.5cm}


\subsection{Component-wise contribution to DW-induced total error}
\label{subsec:componentWise}

Assuming operation at maximum take-off weight, we examine the maximum-throttle case (\(T100\)) in \cref{fig:combined_results}(b,e) relative to \(T0\) means in \cref{fig:IndivPosErrorT0,fig:IndivOriErrorT0}. \cref{fig:combined_results}(b) decomposes the DW-induced \(e^{\mathrm{pos}}\) at \(T100\) into \((e_x,e_y,e_z)\). In all HS, \(e^{\mathrm{pos}}\) is dominated by \(e_x\) and \(e_z\). Outside V5--V6, errors remain \(\approx5\)–\(10~\mathrm{mm}\) in HS1--HS3 and rise to \(\approx25~\mathrm{mm}\) in HS4, with SD of \(\approx5\)–\(15~\mathrm{mm}\) that increases toward HS4. Their sign trends largely follow \cref{fig:IndivPosErrorT0}, with \(e_z\) mostly negative and \(e_x,e_y\) retaining similar azimuth-dependent patterns. By contrast, V5--V6 shows peaks in all components, coincident maxima in \(e^{\mathrm{pos}}\), and larger SD in \(e_x\) and \(e_z\). \cref{fig:combined_results}(e) shows the \(e^{\mathrm{\circ}}\) components at \(T100\). Outside V5--V6, \(e_{\text{yaw}}\) has a negative bias growing from \(\approx15^{\circ}\) in HS1 to \(\approx30^{\circ}\) in HS4, while \(e_{\text{pitch}}\) and \(e_{\text{roll}}\) remain smaller. The largest $e^{\circ}$ peaks are in V5--V8, where all components are perturbed and SD rises from \(<5^{\circ}\) in HS1 to \(\approx30^{\circ}\) in HS4, mainly in \(e_{\text{yaw}}\).



\vspace*{-0.1cm}
\subsection{Evaluating DW effect  under the influence of ground/wall}
\label{subsec:WallANDground}

At \(T100\), \cref{fig:combined_results}(c,f) compare cumulative pose errors in free-space, ground, and wall. Ground effect dominates in HS1, producing the largest \(e^{\mathrm{pos}}\) increase over free-space \(T100\), but this increment decreases with height despite the overall rise in \(e^{\mathrm{pos}}\) with bending. The ground- and wall-effect curves also show lower SD than free-space; the V5--V6 turbulent band remains but with reduced SD. $e^{\circ}$ shows the same trend, with ground and wall reducing mean by \(\approx30\%\) and narrowing SD, keeping the mean near \(10^\circ\) except near the attachment-disturbed region V5--V6.
\vspace{-0.2cm}

\begin{figure}[!t]
    \vspace*{0.1cm}
    \centering
    \includegraphics[width=\linewidth]{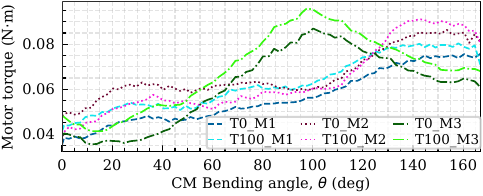}
    \caption{
    Driving motors' (M1,M2$\&$M3) torques when operating in V1, V5 and V9 planes at $T=0\%$ and $T=100\%$.
    }
    \label{fig:MotorTorque}
    \vspace*{-0.6cm}
\end{figure}

\subsection{Effect of DW on driving motor torque of the CM}
Fig.~\ref{fig:MotorTorque} compares the torques of Motors~M1,M2 and M3 for motions in V1, V5, and V9 respectively. These slices align with the tendon--motor symmetry planes about the backbone (\cref{fig:workspace}(a,b)). Hence represent worst-case single-motor winding load. In both T0 and T100, the M1 and M2 peaks occur about \(40^\circ\) later than M3. Across all, the T100 traces are generally above T0, indicating added DW loading. The clearest increase is for M1 and M3: M3’s peak rises from 0.086~Nm to 0.095~Nm (\(+10\%\)), while M1 shows the same increase with a smaller peak. M2 shows the same tendency but with  configuration-dependency and T0--T100 crossover. Overall, DW increases torque across all three motors, though the magnitude and smoothness vary with actuation planes.
\vspace{-0.2cm}



\subsection{CC-guided residual prediction from experimental data}
\label{subsec:GPR_results}

 \begin{table}[h]
\caption{Held-out RMSE of the raw CC prior and the GP-corrected predictor across still-air and DW cases.}
\label{tab:gp_results_all_cases}
\centering
\scriptsize
\resizebox{\columnwidth}{!}{%
\begin{tabular}{lcccc}
\toprule
Case & Position RMSE (mm) & Orientation RMSE ($^\circ$) & Pos. imp. (\%) & Ori. imp. (\%) \\
\midrule
T0   & 124.56 $\rightarrow$ 8.63   & 28.40 $\rightarrow$ 10.86 & 93.00 & 62.42 \\
T25  & 131.52 $\rightarrow$ 14.51  & 24.69 $\rightarrow$ 5.80  & 90.28 & 79.72 \\
T50  & 130.81 $\rightarrow$ 11.94  & 35.20 $\rightarrow$ 18.92 & 91.15 & 47.56 \\
T75  & 132.11 $\rightarrow$ 10.36  & 26.66 $\rightarrow$ 11.29 & 92.17 & 58.64 \\
T100 & 130.90 $\rightarrow$ 14.68  & 24.15 $\rightarrow$ 6.72  & 89.86 & 76.77 \\
\bottomrule
\end{tabular}%
}

\vspace{-1\baselineskip}
\end{table}

\begin{figure}
    \vspace*{0.1cm}
    \setlength{\abovecaptionskip}{-2pt}
    \setlength{\belowcaptionskip}{-8pt}
    \centering
    \includegraphics[width=0.97\linewidth]{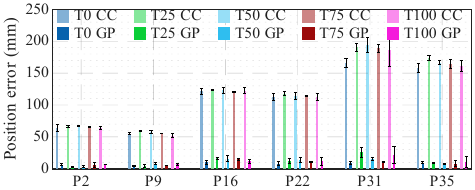}
    
    \caption{Test points' position errors (mean $\pm$ SD) of the raw CC prior and the GP-corrected predictor for \(T0\)–\(T100\).}
    \label{fig:Gaussian_Results_pos}
    \vspace{6pt}
\end{figure}

\begin{figure}
    \setlength{\abovecaptionskip}{-2pt}
    \vspace*{-0.4cm}
    
    \centering
    \includegraphics[width=0.97\linewidth]{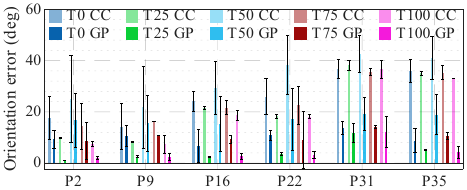}
   \caption{Test points' orientation errors (mean $\pm$ SD) of the raw CC prior and the GP-corrected predictor for \(T0\)–\(T100\).}
    \label{fig:Gaussian_Results_ori}
    \vspace*{-0.5cm}
\end{figure}

Figs.~\ref{fig:Gaussian_Results_pos} and \ref{fig:Gaussian_Results_ori} compare the held-out pointwise errors (mean \(\pm\) SD) of the raw CC prior and GP-corrected predictor for \(T0\)–\(T100\) against experiment. Across all five cases, the GP reduces position error (\(e^{\mathrm{pos}}\)) at all held-out points from \(\sim50\)–\(200\) mm to \(5\)–\(15\) mm. Orientation error is also reduced in every case, though with larger variability, especially under DW, consistent with the stronger free-space fluctuations in \cref{fig:combined_results}(a,d). Position improvement is the most uniform, whereas orientation is more case-dependent: \(T25\) and \(T100\) show the largest gains, while \(T50\) remains the most difficult. Table~\ref{tab:gp_results_all_cases} confirms this trend: position RMSE decreases from \(124.56\text{--}132.11\) mm to \(8.63\text{--}14.68\) mm (\(89.86\%\text{--}93.00\%\)), and orientation RMSE from \(24.15^\circ\text{--}35.20^\circ\) to \(5.80^\circ\text{--}18.92^\circ\), corresponding to (\(47.56\%\text{--}79.72\%\)) improvement.
\vspace{-0.4cm}

%% file: 5-Discussion.tex
\vspace{0.2cm}
\section{DISCUSSION}


\subsection{Evaluating the CM  against its CC model without DW}

\cref{fig:IndivPosErrorT0} shows that the vertical error \(e_z\) is negative and nearly constant within each HS, consistent with gravitational sag and structural effects such as compliance/stiffness that are not captured by the purely geometric CC model \cite{uthayasooriyan2024tendon, amanov2021tendon,5957337}. In contrast, \(e_x\) and \(e_y\) alternate sinusoidally in sign because the sample points lie on circular slices about the CM base, so their projections onto \(x_B\) and \(y_B\) vary with slice angle.

Orientation errors follow the same trend. The scalar metric $e^{\circ}$ captures the net mismatch, and minima at V1/V5/V9 coincide with tendon directions (C1--C3; \cref{fig:workspace}), consistent with the actuation layout. The Euler components are interpreted qualitatively only, since they do not sum to the scalar value. The positive, bounded \(e_{\text{yaw}}\) indicates a repeatable CW bias about \(z_B\), suggesting out-of-plane bending due to fabrication asymmetry and uneven tendon loading; the corrective rotation would therefore be CCW. Likewise, the sign-reversing sinusoidal patterns in \(e_{\text{pitch}}\) and \(e_{\text{roll}}\) are consistent with azimuth-dependent under-tilt about \(y_B\) and \(x_B\), likely related to gravity or backbone twist.


Overall, the negative \(e_z\) and pitch/roll biases indicate reduced effective bending relative to the reference, while \(e_{\text{yaw}}\) reflects out-of-plane bending and influences \(e_x\) and \(e_y\). The mismatch is attributed to both CC geometric idealization and prototype-specific effects, including gravitational sag, compliance/stiffness, torsional deformation, tendon-routing asymmetry, pretension/hysteresis, marker-placement/frame-registration sensitivity, and actuator--pulley transmission sensitivities \cite{picard2018pulleys}. As these effects increase with bending (HS1--HS4), they impose a reduced operational bending range. Higher-fidelity models such as Cosserat rod modelling \cite{webster2010design} could capture effects omitted by the pure CC model.



\subsection{Evaluating the DW effect in free-space}

 DW affects the CM non-uniformly: errors are smallest in HS1, likely because the EE remains closer to the UAV hub and partly outside the stronger wake core, and increase in HS3--4 as the EE moves farther below the vehicle with a larger moment arm. Peaks at V5--V6 beneath Prop.~4 coincide with nearby airframe attachments (Figs.~2 and~3), suggesting a locally disturbed wake and larger CM oscillation, reflected in the wider SD bands. The non-monotonic throttle dependence, with T25 often exceeding T75/T100, may reflect a localized wake--interaction effect rather than a contradiction of stronger overall DW: in hover, multirotor wakes can merge into a non-axisymmetric wake with jets between rotors, and changing rotor speed can alter not only wake strength but also velocity/vorticity distribution and vortex dynamics \cite{wolf2024volumetric,chang2023numerical}. Thus, different throttle settings may change the local impingement pattern on the CM rather than producing a strictly monotonic error increase at every workspace location, motivating future flow-field characterization of CAAMS.

Component analysis shows that DW mainly increases \(e_z\) and \(e_x\), indicating downward push, out-of-plane bending, and twisting reflected in elevated \(e_{\text{pitch}}\), \(e_{\text{roll}}\), and persistent \(e_{\text{yaw}}\). Orientation degrades as the sinusoidal trends in \cref{fig:IndivOriErrorT0} vanish, while the positional pattern broadly follows the benchmark. Peak errors and SD occur in the attachment-disturbed region. Overall, mean error and SD of both Cartesian and Euler components increase with bending, likely due to disturbed-flow-induced oscillations, lever-arm amplification, non-uniform DW, and structural asymmetry. These trends motivate streamlining or repositioning attachments and implementing DW-aware kinematic compensation and control. The bench-top setup was chosen to isolate DW-induced CM deformation from free-flight effects. In flight, wind gusts, UAV-motion-induced drag, attitude-control actions, and coupled CAAMS wrenches may change the magnitude of the deviations, but the qualitative trends remain informative and motivate future in-flight validation with and without control. While Section III-B reports key features for generalization, CM sensitivity is expected to increase for longer, more compliant manipulators and decrease for shorter or stiffer ones, while different multirotor geometries, manipulator dimensions, placements, and morphologies would alter wake overlap and disturbed workspace sectors, motivating broader comparative study and direct throttle-to-RPM calibration or RPM measurement and CM-aerodynamic load distribution modelling for future downwash-informed manipulator control.

\subsection{DW effect under the influence of ground or wall}

Ground amplifies positional error more than wall, especially when the end effector operates near the floor. A possible explanation is that the impinging wake is redirected into a radial wall jet/groundwash with flow separation and recirculating structures near the surface, increasing lateral loading near the distal CM \cite{WU2024204}. By contrast, wall proximity appears weaker because prior multirotor wall-effect experiments indicate a more localized one-sided perturbation of the downwash field than the more global wake redirection caused by the ground \cite{david2024ground}. The reduced SD in the wall/ground cases is therefore interpreted not as weaker disturbance, but as a more surface-constrained and repeatable wake pattern than in free-space T100 \cite{WU2024204,prothin2019aerodynamics,david2024ground}. Further CFD, flow visualization, and speed measurements are needed for a conclusive mechanism.
\vspace{-0.2cm}


\subsection{Effect of DW on CM-motor torque}

Even at T100, the maximum measured torque is 0.095~Nm, only \(\approx6\%\) of the XL430-W250-T stall rating (1.5~Nm at 12~V), leaving enough actuator headroom. The consistent rightward shift of the M1 and M2 peaks at both T0 and T100 likely reflects mismatch in initial tendon tension compared to M3, whereas the irregular crossover in M2 is consistent with the fluctuation-prone V5--V6 region. In heavier and solid surface CM designs, actuator sizing and CAAMS's power budgets should therefore account for DW torque. In non-disturbed regions, the nearly parallel T0 and T100 torque curves suggest that DW acts mainly as an additive load. Thus, if needed, with actuator-specific refinement and tension senors, DW-imposed torque could be incorporated as a feedforward bias term in actuator loading or control design.
\vspace{-0.2cm}


\subsection{CC-guided residual prediction from experimental data}

\cref{subsec:GPR_results} shows that the dominant CC-to-prototype mismatch is learnable. The larger and more uniform gains in position suggest a repeatable translational bias, whereas the smaller orientation gains indicate stronger disturbance-driven variability under DW. Beyond learning-based correction, a more torsionally rigid CM design \cite{peng2023aecom} could further reduce orientation error and increase the usable workspace. Accordingly, the GP should be interpreted as a compact correction layer: \(T0\) provides still-air calibration, while \(T25\)–\(T100\) support DW-aware forward pose prediction for future feedforward compensation and control, rather than validation of the raw CC prior or identification of the aerodynamic load distribution.


%% file: 6-Conclusions.tex
\section{conclusions}


Prior multirotor research shows that understanding DW is essential for both design and control across diverse operating scenarios, including ground/wall/ceiling proximity effects \cite{david2024ground,WU2024204}, aerial spraying \cite{ZHU2022107286}, multi-agent interactions \cite{10804051} and aerial manipulation \cite{Vidyadhara_Tony_Gadde_Jana_Varun_Bhise_Sundaram_Ghose_2022}. In this letter, we experimentally quantified propeller-downwash effects on a tendon-driven continuum manipulator in CAAMS and complemented the study with a CC-guided GPR residual model for still-air calibration and DW-aware forward pose prediction. The still-air \(T0\) case revealed structured prototype--CC mismatch dominated by negative \(e_z\) and yaw-related deviations. Relative to \(T0\), free-space DW primarily perturbs the distal CM, amplifying \(e_z\), and yaw with associated twisting effects. Ground proximity amplifies position error more than wall proximity, while both reduce SD, indicating more repeatable surface-constrained disturbance patterns. Attachments within the wake can strongly degrade stability and accuracy and should therefore be carefully placed. Across held-out points, the GPR model reduced position RMSE to \(8.63\text{--}14.68\) mm (up to 93$\%$) and orientation RMSE to \(5.80^\circ\text{--}18.92^\circ\) (up to 79.72$\%$), substantially improving over the raw CC prior. Torque analysis showed that DW at \(T100\) increases CM driving torque by about \(10\%\) with near parallel shift compared to \(T0\), suggesting a first-order added load that future larger or solid-surfaced-CM CAAMS should account for in actuator sizing and control. In this letter, we focused on experimental disturbance characterization on the EE-pose together with a predictive residual model for DW-aware forward pose estimation. Future work will extend the study across multiple CM and multirotor configurations, integrate CFD and flow-field characterization, and aerodynamic load modelling to produce downwash models for CAAMS, and develop compensation control strategies with in-flight validation.


